\documentclass[11pt, superscriptaddress]{article}
\usepackage{arydshln}
\usepackage{acl2015}
\usepackage{times}
\usepackage{placeins}
\usepackage{graphicx}
\usepackage{mathtools}
\DeclarePairedDelimiter\ceil{\lceil}{\rceil}

\usepackage{url}
\usepackage{algorithm}
\usepackage{algpseudocode}
\usepackage{latexsym}
\usepackage{float}
\makeindex
\title{Building on Huang \textit{et al.}'s GlossBERT for Word Sense Disambiguation}

\usepackage{hyperref}

\author{%
  \textbf{Nikhil Patel, James Hale, Kanika Jindal, Apoorva Sharma,} and \textbf{Yichun Yu}\\
  University of Southern California\\
  \texttt{\{nmpatel, jahale, kjindal, sharmaap, yichunyu\}@usc.edu} \\
}

\begin{document}
\maketitle
\begin{abstract}
We propose to take on the problem of Word Sense Disambiguation (WSD). In language, words of the same form can take different meanings depending on context. While humans easily infer the meaning or gloss of such words by their context, machines stumble on this task. As such, we intend to replicated and expand upon the results of Huang \textit{et al.}'s GlossBERT, a model which they design to disambiguate these words \cite{Gloss}. Specifically, we propose the following augmentations: data-set tweaking ($\alpha$ hyper-parameter), ensemble methods, and replacement of BERT with BART and ALBERT. The following GitHub repository contains all code used in this report, which extends on the code made available by Huang \textit{et al.}: https://github.com/nkhl-p/glossBERT. Additionally, the following links to a short presentation: https://youtu.be/X2OxgcF7lsM.
\end{abstract}

\section{Introduction}
Take the following two sentences: \textit{The dog's \underline{bark} was loud}, and \textit{The tree's \underline{bark} was dark}. In each case, the word \textit{bark} appears with a different sense: the sound a dog makes, and the rough covering of a tree trunk respectively. Note, without context, the disambiguation of \textit{bark} would be impossible; however, with context it becomes trivial for a human, and possible for a computer (the surrounding words \textit{dog} and \textit{tree} would be clues). One can easily see the importance of this, as an NLP model with imprecise WSD may grossly misinterpret a statement: imagine a tree barking, or a dog covered in bark. \par
As another example, consider a sentiment classification model labeling a review containing the word \textit{cheap}; it would be useful for the model to infer whether the reviewer meant \textit{inexpensive} or \textit{poorly made}. Essentially, in this case, glosses intuitively associated with opposite sentiments would be mapped to the same word, which would hamper performance.\par

Lastly, consider a machine translation task in which a model must translate some word $w$ from language $A$ to $B$. An issue arises since the languages $A$ and $B$ do not necessarily share ambiguities; i.e. there may not exist some $w'$ in $B$ such that it shares all glosses with $w$ in $A$ ($w'$ could drop some glosses from $w$). So, it becomes necessary to disambiguate $w$ to ensure we get $w'$ with the proper gloss. To illustrate this, consider the aforementioned \textit{bark} in English; when translating to French, we get \textit{aboient} in the dog-context, and \textit{l'écorce} in the tree-context. \par
One can easily see WSD lies at the foundation of many NLP tasks, and therefore has many traditional NLP applications: e.g. machine translation, information extraction, and sentiment analysis \cite{tang2018analysis,chai1999use,HUNG2016224}. As such, we consider steady improvement in this domain as of great importance, and choose it as our area of focus. \par
To formally define the WSD task, let us reference Huang \textit{et al.} They define a sentence $S$ as a series of words $\{w_0, ..., w_m\}$, and wish to disambiguate some words $w\in S'\subseteq S$; then, each $w_i\in S'$ has some candidate senses, of which the task definition dictates the model select the most appropriate given the context \cite{Gloss}.\par

There are several possible approaches to consider in WSD. In his survey of WSD, Navigli describes three main classes of WSD methods: supervised, where the model generalizes gloss mappings from labeled data; unsupervised, which learn from non-labeled data, and try to disambiguate based on patterns in the training set; and knowledge-based methods, which leverage lexical resources (e.g. dictionaries) to disambiguate words \cite{Navigli}. He further states that supervised systems outperform the other methods \cite{Navigli}. As such, we will focus our efforts specifically in the domain of supervised-WSD. \par

Now, let us introduce the work on which we wish to focus our project. Huang \textit{et al.} propose GlossBERT (a supervised model) for WSD; they feed context-gloss pairs (context being the input sentence, and the glosses coming from WordNet) into a BERT-based binary classifier for the disambiguation task, where the model will determine whether the context fits the gloss \cite{Gloss}. Specifically, for each ambiguous word $w$ in a sentence, the authors create $N$ context-gloss pairs with different senses of $w$ and a label corresponding with each pair (positive if the gloss matches, negative if not) \cite{Gloss}. Further, the authors propose three variations of GlossBERT: Token-CLS, Sent-CLS, and Sent-CLS-WS \cite{Gloss}. \par
\par% \textbf{explain models a little here} In the Sent-CLS-WS model, the authors augment the context-gloss pairs with weak-supervised signals to aid the model in identifying the target word from input \cite{Gloss}.\par
Principally, we propose to replicate and expand on GlossBERT \cite{Gloss}. Initially, after replication of results, we will tweak hyper-parameters to either justify or contradict the authors' choices. Lastly, we intend to conduct a more invasive series of tests involving GlossBERT's classification architecture---e.g. attempting ensemble methods, and replacing BERT with BART and ALBERT. \par

\section{Methods and Experiments}
\subsection{Data-sets}
% Huang \textit{et al.} test their model on only English WSD data-sets \cite{Gloss}. We believe it important to test model performance on languages with a range of syntactic structures. As such, we intend to not only test on the data-sets from the original paper, but also on non-English WSD data-sets such as ItalWordNet:Senseval-3 task 02, Italian words corpus.\par
% \textbf{Training corpus:} We choose SemCor3.0 as our training corpus since it is the ``largest corpus manually annotated with WordNet sense for WSD'' \cite{Gloss}. We use the current techniques described in the paper under consideration to maintain our metricts compatible with \cite{Gloss} and to evaluate our techniques.\par
\textbf{Training Datasset:} We choose the smallest data-set, Senseval-2, among all test sets used in the Senseval and SemEval context for training, merely due to its size \cite{edmonds-cotton-2001-senseval}. Initially, we tried training on SemCor, as did Huang \textit{et al.}, however we were unable to accommodate such a large training set with memory and time constraints. Therefore, we train all our models on Senseval-2 (SE2).

\textbf{Evaluation Dataset:} For comparable results we use the benchmark datasets proposed in GlossBERT \cite{Gloss}, employing the standard all-words fine-grained WSD datasets from the Senseval and SemEval competition namely:

\begin{itemize}
% \item \textbf{Senseval-2} (SE2) \cite{edmonds-cotton-2001-senseval}
\item \textbf{Senseval-3} (SE3) \cite{mihalcea-etal-2004-senseval}
\item \textbf{SemEval-2007} (SE07) Task 17 \cite{pradhan-etal-2007-semeval}
\item \textbf{SemEval-2013} (SE13) Task 12 \cite{navigli-etal-2013-semeval}
\item \textbf{SemEval-2015} (SE15) Task 13 \cite{moro-navigli-2015-semeval}
\end{itemize}

\textbf{Dataset Statistics:} The statistics of the WSD datasets are presented in GlossBERT \cite{Gloss} in Table~\ref{fig:set_stats}.

\begin{table}
\begin{small}
\label{pos-stats}
\begin{tabular}{|c|c c c c c|}
\hline
Data-set & Total & Noun & Verb & Adj & Adv \\
\hline
SemCor & 226036 & 87002 & 88334 & 31753 & 18947 \\
% \hdashline
SE2 & 2282 & 1066 & 517 & 445 & 254 \\
% \hdashline
SE3 & 1850 & 900 & 588 & 350 & 12 \\
% \hdashline
SE07 & 455 & 159 & 296 & 0 & 0 \\
% \hdashline
SE13 & 1644 & 1644 & 0 & 0 & 0 \\
% \hdashline
SE15 & 1022 & 531 & 251 & 160 & 80 \\
\hline
\end{tabular}
\caption{Part of speech tags in WSD dataset (table from Huang \textit{et al.}) \protect\cite{Gloss}}
\label{fig:set_stats}

\end{small}
\end{table}

\subsection{Trials}
\textbf{Base}: Technical limitations disallowed complete replication of Huang \textit{et al.}'s results. However, we compromised and trained on a smaller data-set (Senseval-2) with a smaller batch-size (10). Obviously, this led to worse performance than that from the original paper. However training the original models in this manner gave us a baseline to which we could compare augmentations. \par
\textbf{Making an $\alpha$ hyper-parameter:} Intuitively, the more possible labels of an instance, the longer (or more iterations) it will take to learn; conversely, the less labels, the shorter. We draw from this intuition to reform the data-set to embody this idea---i.e. ambiguous words with more glosses will appear in the data-set disproportionately more. More formally, let $N$ be the number of glosses of an ambiguous word and $\alpha$ be some hyper-parameter. Then, we randomly sample $N^\alpha-1$ data-points from the $N$ data-points described by Huang \textit{et al.} (binary classification points for context-gloss pairs, with a positive label for the correct pair), and add the positive pair to ensure it appears at least once in the training-set \cite{Gloss}. As such, given a word $w$, its set of glosses $G_w$, and a hyper-parameter $\alpha$, we use Algorithm~\ref{alg:N} to construct some data-points to append to our training-set. \par

% \textbf{Making $N$ a hyper-parameter:} In the versions of GlossBERT presented by Huang \textit{et al.}, when generating data-sets for the models, the number of data-points for a word appear proportionally to how many glosses WordNet has for it \cite{Gloss}. However, intuition implies a model may need to see a data-point disproportionately more times for learning. As such, given a word $w$, its set of glosses $G_w$, and a hyper-parameter $\alpha$, we use Algorithm~\ref{alg:N} to construct some data-points to append to our data-set. 
\begin{algorithm}
\caption{Constructs array of data-points append when constructing train-set (note `$+$' means concatenation of two arrays)}\label{alg:N}
\begin{algorithmic}
% $y \gets 1$\;
% \Require $n \geq 0$
% \Ensure $y = x^n$
\Procedure{Train-Gloss}{$w,\alpha$}      
% \State $\alpha \gets$ Hyper-parameter defined beforehand
% \State $w \gets$ Word to disambiguate
\State $G_w \gets$ Set of glosses of $w$ from WordNet
\State $G_w' \gets$ Empty array
\State $g\gets x$ s.t. $x\in G_w$ labeled positive
\For{\texttt{ $0$ to $\ceil*{|G_w|^{\alpha}}$}}
        \State $x \stackrel{R}{\gets} G_w$ \Comment{s.t. $x$ is a uniformly sampled gloss from $G_w$}
        \State $G_w' \gets G_w' +$ arrray($x$) \Comment{Add newly sampled point to $G_w'$}
      \EndFor
     \State $G_w' \gets G_w' +$ arrray($g$) \Comment{Ensure at least one positive labeled point in $G_w'$}
% \While{$N \neq 0$}
% \If{$N$ is even}
%     \State $X \gets X \times X$
%     \State $N \gets \frac{N}{2}$  \Comment{This is a comment}
% \ElsIf{$N$ is odd}
%     \State $y \gets y \times X$
%     \State $N \gets N - 1$
% \EndIf
% \EndWhile
\EndProcedure
\end{algorithmic}
\end{algorithm}

\begin{figure}[tb]
   \begin{minipage}{0.5\textwidth}
     \centering
\includegraphics[width=\columnwidth]{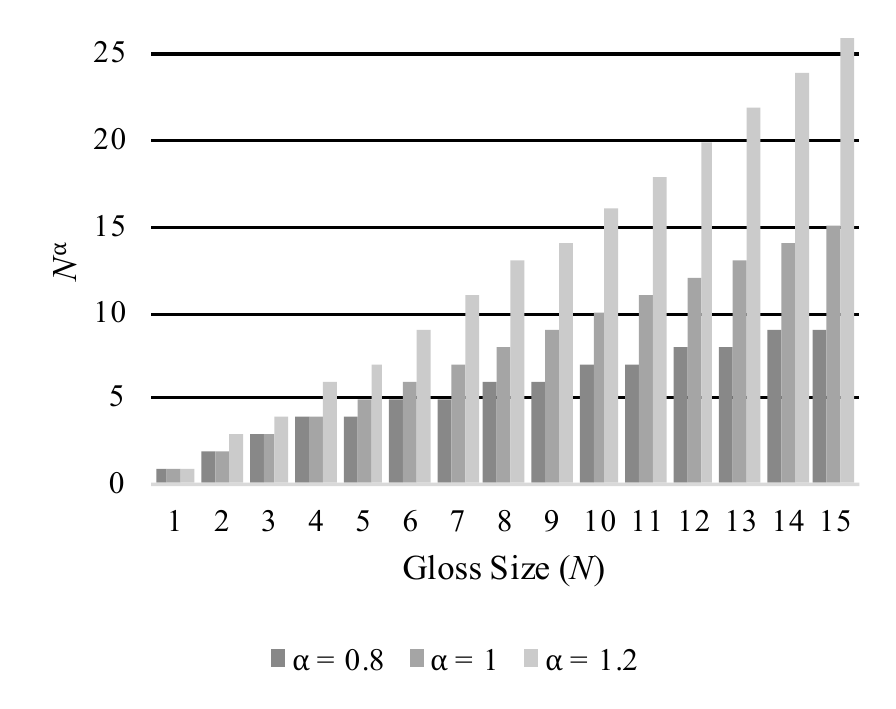}
\end{minipage}
\caption{Demonstrates how number of points for ambiguous word (with $N$ possible glosses) changes with different $\alpha$ and $N$ values}
\label{fig:N}
\end{figure}

\begin{figure}[tb]
   \begin{minipage}{0.4\textwidth}
     \centering
\includegraphics[width=\columnwidth]{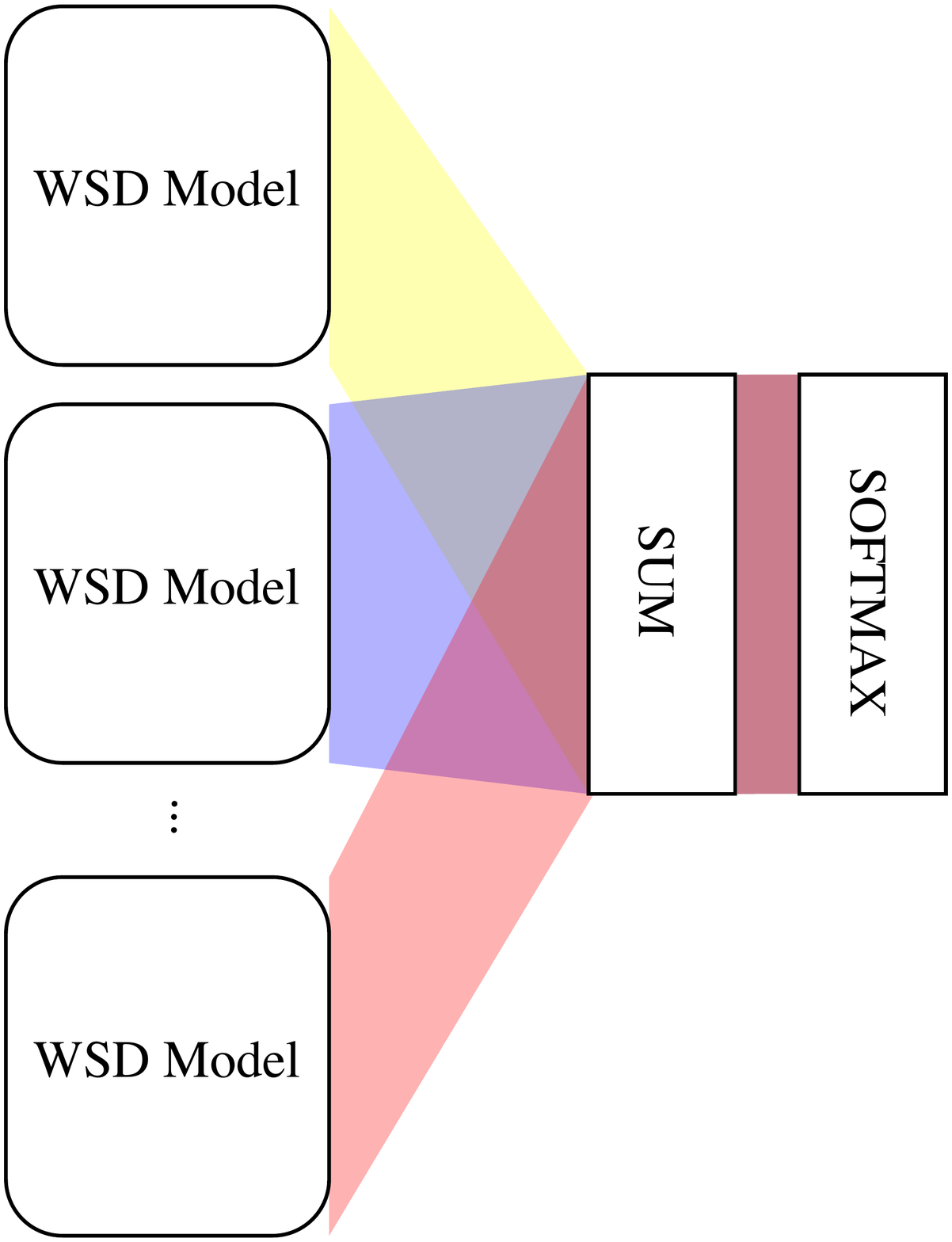}
\end{minipage}
\caption{Ensemble diagram}
\label{fig:ensemble}
\end{figure}

\textbf{Ensemble Methods: } Ensemble methods allow for combination of various models in the hopes of improving performance; they have proven themselves quite effective in myriad scenarios and forms (e.g. random forests). As such, we investigated whether an ensemble of WSD models could improve performance. In the proceeding trials, we used a heterogeneous combination of model types---taking our trained models from the base-case, and with various $\alpha$ values, and combining them in a manner demonstrated in Figure~\ref{fig:ensemble}. Unfortunately, the BERT(Token-CLS) model had a different tensor output shape than the other three models, so we did not include it in any ensemble trials; specifically, we conducted ensemble trials with a combination of GlossBERT(Sent-CLS-WS), GlossBERT(Token-CLS), and GlossBERT(Sent-CLS).  Essentially, we take their outputs before the final softmax, add them, and pass the resulting tensor through softmax activation. 

\textbf{Replacing BERT with ALBERT}: Another intuitive approach to improve the existing approach is to replace the BERT model with a better version of it, ALBERT. After studying multiple existing researches on BERT, we found that ALBERT is a much better version of BERT, as it reduces the memory consumption of the model and simultaneously increasing the speed by significant amount.

Per the huggingface documentation, the key idea behind this improvement is 
\begin{itemize}
    \item (i) ``splitting the embedding matrix into two smaller matrices'' \cite{huggingfaceAl}
    \item (ii) ``repeating layers split among groups'' \cite{huggingfaceAl}
\end{itemize}
 
% ALBERT model uses self-supervised loss that helps modeling inter-sentence coherence.

ALBERT model is computationally the same as BERT model as it iterates through same number of layers and repeating layers result in smaller memory consumption \cite{huggingfaceAl}.
% The performance improvement of albert can be attributed to the following reasons: 
% \begin{itemize}
%     \item (i) Factorizing the embedding parameters, decouples them from the hidden state, thus preventing large size embeddings. Thereby, reducing the number of model parameters.
%     \item (ii) In encoder and feedforward layer, parameter sharing between the attention sub-segments within a layer, is referred to as Cross-layer parameter. This also reduces the number of parameters significantly.
% \end{itemize}
Our approach is to replace BERT model with ALBERT model. The following are the conclusions and challenges faced:
\begin{itemize}
    \item \textbf{Similar class design architecture}: \newline AlbertConfig :: BertConfig, \newline AlbertForPreTrained :: BertForZreTrained.
    \item  \textbf{Different input parameters, output values and class methods}: Authors of GlossBERT rewrite large amounts of code to build GlossBERT.
    % \item Memory consumption was reduced and training speed improved but \textbf{the result has not been improved}, as show in Table 2.

\end{itemize}

\textbf{Replacing BERT with BART:} Another intuitive approach to improve the existing approach is to replace the BERT model with BART.
% , a denoising autoencoder for pretraining sequence-to-sequence models. 
% BART is particularly effective when fine tuned for text generation but also works well for comprehension tasks.
We chose BART as the model that can change the learning approach to discriminative.
In the paper \cite{BART} the author mentions that on experimenting with BART, the model does not work well on WSD. On trying techniques to replace BERT completely with BART we confirmed that the findings of author of \cite{BART} that existing trained BART model is not suitable for word sense disambiguation tasks. 
Even after adding parameters required for BART, which are more than BERT, the findings are similar and we found that classes and functions given in the \cite{huggingfaceBa}, BART and BERT are very different and hence make BART not compatible for changes in the GlossBERT.
In the \cite{BART}, the author has provided with following results:

\begin{table}[H]
\begin{small}
\label{pos-stats}
\begin{tabular}{|c|c c c c c|}
\hline
System & SE7 & SE2 & SE3 & SE13 & SE15 \\
\hline
BERT(2019) & 71.99 & 77.8 & 74.6 & 76.5 & 79.7\\
% \hdashline
BART(2020) & 67.2 & 77.6 & 73.1 & 77.5 & 79.7\\
% \hdashline
Albert(2020) & 71.4 & 75.9 & 73.9 & 76.8 & 78.8\\
\hline
\end{tabular}
\caption{SemCor test results of LMGC for base trained former models (table from \protect\cite{BART})}
\end{small}
\end{table}
showing that BERT is at par with the latest models such as BART.
\section{Results and Discussion}
Table \ref{fig:scoretable} shows the performance of Huang \textit{et al.}'s models \cite{Gloss} in comparison to  ours. We were able to get the following results by setting the hyper-parameters outlined in Figure~\ref{fig:parameters} (unless otherwise denoted). Of note, Sent-CLS-WS, Token-CLS, and Sent-CLS models performed better over Base BERT Token-CLS model. A proportionately similar trend was observed from the results of previously published work \cite{Gloss}. Accommodating the batch in GPU memory for training presented as a major challenge, which caused us to limit ourselves to a batch size of 10 in comparison to 64 used in the original paper \cite{Gloss}. We encountered over-fitting as another major obstacle. We tried experimenting with different optimizers and learning rates, but were unable to solve this problem due to GPU memory limits (16 GB on Google Colab Pro). To further optimize our results with these limitations, we employed gradient accumulation---set to 3---to help prevent over fitting, resulting in slight improvement in comparison to updating gradients on every batch.\par
We ran two experiments varying the aforementioned $\alpha$ hyper-parameter. In the first experiment, we set $\alpha$ to 0.8 in hopes of mitigating over-fitting. In the second, we set $\alpha$ to 1.2 to reinforce the learning based in the number of glosses of a word in WordNet. While our attempt to avoid over-fitting didn't turn out to be effective, it helped in improving overall accuracy marginally across other development data sets, which can be seen in Table~\ref{fig:scoretable}. \par
Unfortunately, the ensemble methods under-performed, unable to defeat each of their components (though defeating some of them). For improvement, in future work, we would like to investigate homogeneous combinations of ensemble methods with popular ensemble techniques---e.g. bagging; we would also like to experiment with various voting schemes, rather than just combining output as done here; and lastly, we would like to increase the number of models in the ensemble (here we only have three).
%  In an attempt to improve the model by disproportionate gloss pairs, and to elevate the issue of over fitting we devised two experiments with the hyper parameter. \textbf{Experiment 1:} We set $\alpha$ to 0.8, as an attempt to  avoid over fitting and Experiment 2: setting $\alpha$ to 1.2, as an attempt to reinforce the learning based in the number of glosses WordNet has. Even though, our attempt to avoid over fitting didn't turn out to be effective, however it helped in improving overall accuracy marginally across other development data sets, which can be seen in Table \ref{fig:scoretable}.

\begin{figure}[h]
    \centering
\begin{tabular}{ |c | c   c| }
\hline
 & \textbf{Value} &\\
\textbf{Variables} & \small{Us} & \small{Paper}\\
\hline
 {\fontfamily{qcr}\selectfont max\_seq\_length} & 512  & 512\\ 
  {\fontfamily{qcr}\selectfont train\_batch\_size} & \textbf{10} & \textbf{64} \\ 
    % {\fontfamily{qcr}\selectfont eval\_batch\_size} & 128  & 128\\ 
      {\fontfamily{qcr}\selectfont learning\_rate} & 2e-6 & 2e-5 \\ 
  {\fontfamily{qcr}\selectfont num\_train\_epochs} & \textbf{6.0}  & \textbf{4.0}\\ 
  Training set & \textbf{SE02}  & \textbf{SE07}\\ 
%   Testing set & SE13 & SE13${}^*$\\
 \hline
\end{tabular}
    \caption{Hyper-parameters and other variables used for results in Figure~\ref{fig:scoretable} from us and Huang \textit{et al.} with differences bolded (${}^*$note Huang \textit{et al.} tested on more data-sets, but we used their results from SE13 for comparison) \protect\cite{Gloss}}
    \label{fig:parameters}
\end{figure}

\begin{table*}[!tbh]
    \centering
    \begin{tabular}{|c|l|c|c|c|c|}
    \hline
        \textbf{Data-set} & \textbf{Model} & \textbf{Huang \textit{et al.}} & \textbf{Base} & \textbf{Exp 1} ($\alpha$=0.8) & \textbf{Exp 2} ($\alpha$=1.2) \\ 
        \hline
        % ~ SE02(dev)  & GlossBERT(Sent-CLS-WS)    & \textbf{77.7} & 96.7 & 88.1 & 94.7     \\
        % ~       & GlossBERT(Token-CLS)      & 77.0 & \textbf{99.3} & \textbf{90.6} & \textbf{96.2}     \\ 
        % ~       & GlossBERT(Sent-CLS)       & 76.5 & 97.1 & 88.2 & 95.6     \\ 
        % ~       & BERT(Token-CLS)           & 69.7 & 80.7 & 76.0 & 76.0     \\ \hdashline
        ~ SE03  & GlossBERT(Sent-CLS-WS)    & 75.2 & 44.3 & 51.1 & 52.4     \\ 
        ~       & GlossBERT(Token-CLS)      & \textbf{75.4} & 48.5 & 50.9 & 52.9     \\ 
        ~       & GlossBERT(Sent-CLS)       & 73.4 & \textbf{51.9} & \textbf{60.9} & \textbf{60.9}     \\ 
        ~       & BERT(Token-CLS)           & 69.4 & 32.4 & 32.9 & 32.9     \\ 
        & Ensemble &N/A &49.4 &49.9& 51.5 \\\hdashline
        ~ SE07  & GlossBERT(Sent-CLS-WS)    & \textbf{72.5} & 39.8 & 44.0 & 42.6     \\ 
        ~       & GlossBERT(Token-CLS)      & 71.9 & \textbf{45.9} & 45.3 & 46.4     \\ 
        ~       & GlossBERT(Sent-CLS)       & 69.2 & 43.3 & \textbf{57.6} & \textbf{56.3}     \\ 
        ~       & BERT(Token-CLS)           & 61.1 & 24.6 & 25.9 & 25.9     \\ 
                & Ensemble &N/A &42.6&41.1&43.3 \\\hdashline

        ~ SE13  & GlossBERT(Sent-CLS-WS)    & \textbf{76.1} & 60.3 & 56.3 & 55.9     \\ 
        ~       & GlossBERT(Token-CLS)      & 74.6 & 61.4 & 56.9 & 58.4     \\ 
        ~       & GlossBERT(Sent-CLS)       & 75.1 & \textbf{64.6} & \textbf{72.3} & \textbf{72.7}     \\ 
        ~       & BERT(Token-CLS)           & 65.8 & 41.5 & 42.2 & 42.2     \\ 
            & Ensemble &N/A &65.1&63.9&65.1 \\\hdashline
        ~ SE15  & GlossBERT(Sent-CLS-WS)    & \textbf{80.4} & 59.4 & 59.7 & 61.0     \\ 
        ~       & GlossBERT(Token-CLS)      & 79.5 & 66.6 & 59.8 & 60.6     \\ 
        ~       & GlossBERT(Sent-CLS)       & 79.5 & \textbf{68.5} & \textbf{71.0} & \textbf{73.0}     \\ 
        ~       & BERT(Token-CLS)           & 69.5 & 41.8 & 59.8 & 59.8     \\ 
                & Ensemble &N/A &65.2&64.9& 64.4\\\hline
    \end{tabular}
     \caption{Results of the four models using Huang \textit{et al.}'s implementation of GlossBERT, and comparing to their results (F1 score as metric) as described in their paper \protect\cite{Gloss}}
    \label{fig:scoretable}
\end{table*}

\begin{figure*}[tbh!]
   \begin{minipage}{\textwidth}
     \centering
\includegraphics[width=\columnwidth]{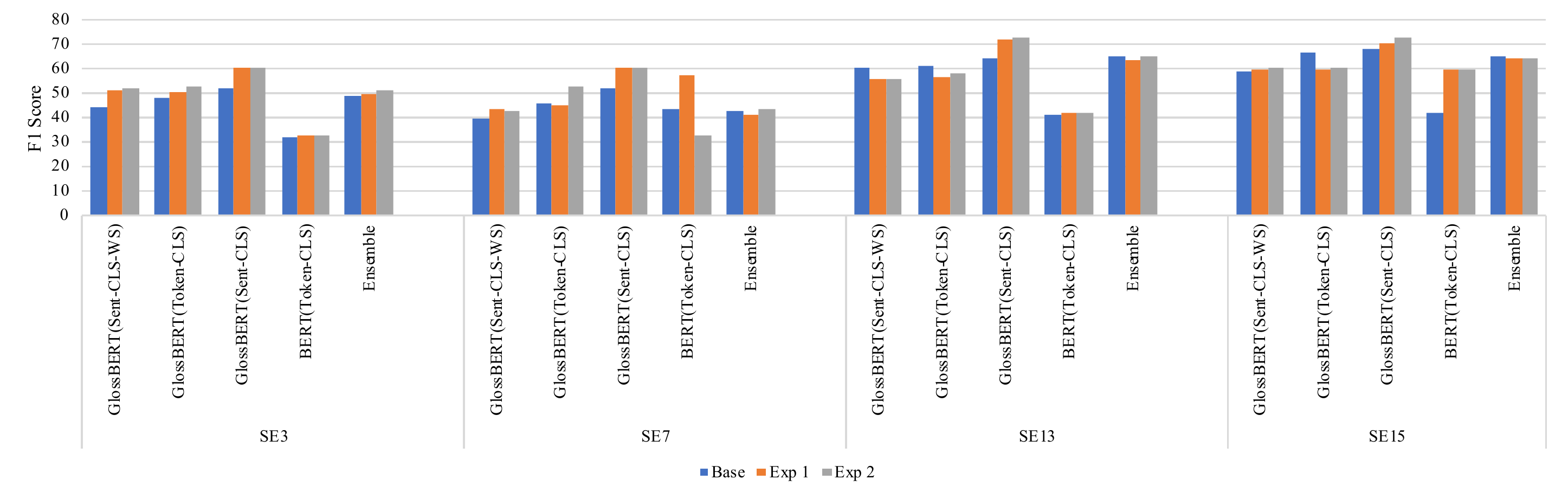}
\end{minipage}
\caption{Visualization of Table~\ref{fig:scoretable}}
\label{fig:N}
\end{figure*}

\section{Conclusion}
In this paper, we investigated Huang \textit{et al.}'s GlossBERT and attempted to expand on their work. We saw some success against our baseline with our $\alpha$ hyper-parameter. However, the ensemble methods could not outperform all of their components. We listed possible future work with more ensemble techniques, and also we could try to follow through with the BART and ALBERT directions. Lastly, we would be curious to see whether these results hold up when training on SemCor with higher compute power.

\section*{Acknowledgement}
First, we would like to thank Huang \textit{et al.}, authors of \textit{Glossbert: BERT for word sense disambiguation with gloss knowledge}, for making their code publicly available. We produced all presented results from their code, or a slightly modified version fitting our needs. \par
We would also like to thank the Professors and TAs of CSCI-544 for their thoughtful advice and assistance.
\bibliographystyle{acl}
\bibliography{Proposal.bib}

% \begin{thebibliography}{}

% \bibitem[\protect\citename{Aho and Ullman}1972]{Aho:72}
% Alfred~V. Aho and Jeffrey~D. Ullman.
% \newblock 1972.
% \newblock {\em The Theory of Parsing, Translation and Compiling}, volume~1.
% \newblock Prentice-{Hall}, Englewood Cliffs, NJ.

% \bibitem[\protect\citename{{American Psychological Association}}1983]{APA:83}
% {American Psychological Association}.
% \newblock 1983.
% \newblock {\em Publications Manual}.
% \newblock American Psychological Association, Washington, DC.

% \bibitem[\protect\citename{{Association for Computing Machinery}}1983]{ACM:83}
% {Association for Computing Machinery}.
% \newblock 1983.
% \newblock {\em Computing Reviews}, 24(11):503--512.

% \bibitem[\protect\citename{Chandra \bgroup et al.\egroup }1981]{Chandra:81}
% Ashok~K. Chandra, Dexter~C. Kozen, and Larry~J. Stockmeyer.
% \newblock 1981.
% \newblock Alternation.
% \newblock {\em Journal of the Association for Computing Machinery},
%   28(1):114--133.

% \bibitem[\protect\citename{Gusfield}1997]{Gusfield:97}
% Dan Gusfield.
% \newblock 1997.
% \newblock {\em Algorithms on Strings, Trees and Sequences}.
% \newblock Cambridge University Press, Cambridge, UK.

% \end{thebibliography}

\end{document}